# Fast Restricted Causal Inference

*Mieczysław A. Kłopotek*

*Institute of Computer Science, Polish Academy of Sciences*

*e-mail: klopotek@ipipan.waw.pl*

**Abstract**

Hidden variables are well known sources of disturbance when recovering belief networks from data based only on measurable variables. Hence models assuming existence of hidden variables are under development. This paper presents a new algorithm "accelerating" the known CI algorithm of Spirtes, Glymour and Scheines [20]. We prove that this algorithm does not produces (conditional) independencies not present in the data if statistical independence test is reliable. This result is to be considered as non-trivial since e.g. the same claim fails to be true for FCI algorithm, another "accelerator" of CI, developed in [20].

**Keywords:** Belief networks, discovery under causal insufficiency,

# 1  Introduction

It is a well known phenomenon of human mind to think in terms of causality. The background behind this paradigm is a strong belief that an event may in fact have only few causes so that reasoning about real world events may be kept from explosion of alternative explanations by identifying intrinsic causality. Even in the domain of stochastic relationships the paradigm of causality proved to be quite helpful.

Belief networks, bayesian networks, causal networks, or influence diagrams, or (in Polish) cause-effect networks are terms frequently used interchangeably. They are quite popular for expressing causal relations under multiple variable setting both for deterministic and non-deterministic (e.g. stochastic) relationships in various domains: statistics, philosophy, artificial intelligence [4], [18].

Various expert systems, dealing with uncertain data and knowledge, possess knowledge representation in terms of a belief network (e.g. knowledge base of the MUNIM system[1], ALARM network [3] etc.). A number of efficient algorithms for propagation of uncertainty within belief networks and their derivatives have been developed, e.g. [14], [16], [17] and many other.

Though a belief network (a representation of the joint probability distri-bution, see [4]) and a causal network (a representation of causal relationships [18]) are intended to mean different things, they are closely related. Both assume an underlying DAG (directed acyclic graph) structure of relations among variables and if Markov condition and faithfulness condition [20] are met, then a causal network is in fact a belief network. The difference comes to appearance when we recover belief network and causal network structure from data. A DAG of a belief network is satisfactory if the generated probability distribution fits the data, may be some sort of minimality is required. A causal network structure may be impossible to recover completely from data as not all directions of causal links may be uniquely determined [20]. Fortunately, if we deal with causally sufficient sets of variables (that is whenever significant influence variables are not omitted from observation), then there exists the possibility to identify the family of belief networks a causal network belongs to [21].

A similar result is harder to establish for causally insufficient sets of variables (that is when significant influence variables are hidden) - "Statistical indistinguishability is less well understood when graphs can contain variables representing unmeasured common causes" ([20], p. 88). Latent (hidden) variable identification has been investigated intensely both for belief networks (e.g. [13], [7], [12], [3]) and causal networks ([15], [18], [20], [5], [6]), and not only in traditional statistics ([8], [2]), [19]). A relationship between the causal network resulting from Causal Inference (CI) algorithm (from chapter 6. of [20]) and a family of belief networks has been established in [11].

Though the CI algorithm possesses interesting properties from the point of view of establishing causal relationships, prediction ([20], Chapters 6-10) as well as from the point of view of belief network construction (see e.g. [11]), it has a major drawback: it is feasible only for small networks (of less nodes than 10). For this reason, another algorithm - FCI - has been developed in [20], Chapter 6, which is claimed feasible for networks of several dozens of nodes. Regrettably, it has been proven erroneous for networks of 30 nodes and more (with 70 links and more) with special structure - compare [10].Another algorithm with latent, IC from [15], is known to be wrong for networks of less than 10 nodes - compare Discussion in chapter 6 of [20].

This paper presents another algorithm, called here Fr(k)CI - a derivative of CI - for construction of a belief network when hidden variables are to be expected. This algorithm "accelerates" CI in that conditional independence is checked only on up to k variables instead of all variables as required by CI. This reduces the amount of data necessary to establish reliable conditional independence. This happens, at the expense of adding superfluous causal links. Their number, however, should not be too high if the intrinsic causal network is not too dense. FCI also added superfluous causal links during its run and this was demonstrated to be the cause of errors in some cases [10]. Therefore the proof of correctness of the Fr(k)CI algorithm is presented in the Appendix A.

## 2 Basic Ideas Beyond Spirtes et al. CI Algorithm

Hidden (latent) variables are source of trouble both for identification of causal relationships (well-known confounding effects) and for construction of a belief network (ill-recognized direction of causal influence may lead to assumption of independence of variables not present in the real distribution). Hence much research has been devoted to construction of models with hidden variables. It is a trivial task to construct a belief network with hidden variables correctly reflecting the measured joint distribution. One can consider a single hidden variable upon which all the measurables depend on. But such a model would neither meet the requirements put on belief network (space saving representation of distribution, efficient computation of marginals and conditionals) nor those for causal networks (prediction capability under control of some variables). Therefore, criteria like minimal latent model [15] or maximally informative partially oriented path graph [20] have been proposed. As the IC algorithm for learning minimal latent model [15] is known to be wrong, let us consider the CI algorithm from [20].

In [20] the concept of including path graph is introduced and studied. Given a directed acyclic graph G with the set of hidden nodes $V_h$ and visible nodes $V_s$ representing a causal network CN, an including path between nodes A and B belonging to $V_s$ is a path in the graph G such that the only visible nodes (except for A and B) on the path are those where edges of the path meet head-to-head and there exists a directed path in G from such a node to either A or B. An including path graph for G is such a graph over $V_s$ in which if nodes A and B are connected by an including path in G ingoing into A and B, then A and B are connected by a bidirectional edge $A<->B$. Otherwise if they are connected by an including path in G outgoing from A and ingoing into B then A and B are connected by an unidirectional edge $A->B$. As the set $V_h$ is generally unknown, the including path graph (IPG) for G is the best we can ever know about G. However, given an empirical distribution (a sample), though we may be able to detect presence/absence of edges from IPG, we may fail to decide uniquely orientation of all edges in IPG. Therefore, the concept of a partial including path graph was considered in [20]. A partially oriented including path graph contains the following types of edges unidirectional: $A->B$, bidirectional $A<->B$, partially oriented $Ao->B$ and non-oriented $Ao-oB$, as well as some local constraint information $A*-\underline{*B*}-*C$ meaning that edges between A and B and between B and C cannot meet head to head at B. (Subsequently an asterisk (*) means any orientation of an edge end: e.g. $A*->B$ means either $A->B$ or $Ao->B$ or $A<->B$). A partial including path graph (PIPG) would be maximally informative if all definite edge orientations in it (e.g. $A-*B$ or $A<-*B$ at A) would be shared by all candidate IPG for the given sample and vice versa (shared definite orientations in candidate IPG also present in maximally informative PIPG), the same should hold for local constraints. Recovery of the maximally informative PIPG is considered in [20] as too ambitious and a less ambitious algorithm CI has been developed therein producing a PIPG where only a subset of edge end orientations of the maximally informative PIPG are recovered. Authors of CI claim such an output to be still useful when considering direct and indirect causal influence among visible variables as well as some prediction tasks.

We cite below some useful definitions from [20].

In a partially oriented including path graph $\pi$:

(i) A is a parent of B if and only if edge $A-> B$ is in $\pi$.

(ii) B is a collider along the path $<A, B, C>$ if and only if $A * -> B <-* C$ in $\pi$.

(iii) An edge between B and A is into A iff $A <-* B$ is in $\pi$

(iv) An edge between B and A is out of A iff $A-> B$ is in $\pi$.

(v) In a partially oriented including path graph $\pi$, U is a definite discriminating path for B if and only if U is an undirected path between X and Y containing B, $B \neq X, B \neq Y$, every vertex on U except for B and the endpoints is a collider or a definite non-collider on U and:

(a) if V and V" are adjacent on U, and V" is between V and B on U, then $V *-> V"$ on U,

(b) if V is between X and B on U and V is a collider on U, then $V-> Y$ in $\pi$, else $V <-* Y$ on $\pi$

(c) if V is between Y and B on U and V is a collider on U, then $V-> X$

in $\pi$, else $V <-*X$ on $\pi$

(d) X and Y are not adjacent in $\pi$.

(e) Directed path U: from X to Y: if V is adjacent to X on U then $X->V$ in $\pi$, if $V$ is adjacent to Y on V, then $V->Y$, if V and V" are adjacent on U and V is between X and V" on U, then $V->V$" in $\pi$.

## 3 The New Algorithm

Let us introduce some notions specific for Fr(k)CI:

(i) A is r(k)-separated from B given set S ($card(S) \le k$) iff A and B are conditionally independent given S

(ii) In a partially oriented including path graph $\pi$, a node A is called *legally removable* iff there exists no local constraint information $B*-\underline{*A*}-*C$ for any nodes B and C and there exists no edge of the form $A*->B$ for any node B.

**The Fast Restricted-to-k-Variables Causal Inference Algorithm (Fr(k)CI):**

Input: Empirical joint probability distribution

Output: Belief network.

**A)** Form the complete undirected graph Q on the vertex set V.

**B')** for j=0 step 1 to k

do if A and B are r(k)-separated given any subset S of neighbours of A or of B, card(S)=j, remove the edge between A and B, and record S in Sepset(A,B) and Sepset(B,A).

**B")** if A and B are r(k)-separated given any subset S of V ($card(S) > 0$), remove the edge between A and B, and record S in Sepset(A,B) and Sepset(B,A).

**C)** Let F be the graph resulting from step B). Orient each edge as $o - o$ (unoriented at both ends). For each triple of vertices A,B,C such that the pair A,B and the pair B,C are each adjacent in F, but the pair A,C are not adjacent in F, orient A\*-\*B\*-\*C as $A * - > B < - * C$ if and only if B is not in Sepset(A,C), and orient A\*-\*B\*-\*C as $A * - \underline{*B*} - *C$ if and only if B is in Sepset(A,C).

**D)** Repeat

(**D1**) **if** there is a directed path from A to B, and an edge A\*-\*B, orient A\*-\*B as $A * - > B$,

(**D2**) **else if** B is a collider along $< A, B, C >$ in $\pi$, B is adjacent to D, A and C are not adjacent, and there exists local constraint $A * - \underline{*D*} - *C$, then orient $B * - * D$ as $B < - * D$,

(**D4**) **else if** $P * - \underline{> M*} - *R$ then orient as $P * - > M - > R$.

(**D3**) **else if** U is a definite discriminating path between A and B for

> M in $\pi$ and P and R are adjacent to M on U, and P-M-R is a triangle, then
>
> if M is in Sepset(A,B) then M is marked as non-collider on subpath $P*-\underline{*M*}-R$
>
> else $P*-*AM*-*R$ is oriented as $P*->M<-*R$,

**until** no more edges can be oriented.

**E)** Orient every edge $Ao->B$ as $A->B$.

**F)** Copy the partially oriented including path graph $\pi$ onto $\pi'$.

Repeat:

In $\pi'$ identify a legally removable node A. Remove it from $\pi'$ together with every edge $A*-*B$ and every constraint with A involved in it. Whenever an edge $Ao-oB$ is removed from $\pi'$, orient edge $Ao-oB$ in $\pi$ as $A<-B$.

Until no more node is left in $\pi'$.

**G)** Remove every bidirectional edge $A<->B$ and insert instead parentless hidden variable $H_{AB}$ adding edges $A<-H_{AB}->B$

**End of Fr(k)CI**

# 4  Differences to Spirtes et al. CI Algorithm

Steps E) and F) constitute an extension of the original CI algorithm of [20], bridging the gap between partial including path graph and the belief network.

Step B) was modified by substituting the term "d-separation" with "r(k)-separation". This means that not all possible subsets S of the set of all nodes V (with card(S) up to card(V)-2) are tested on rendering nodes A and B independent, but only those with cardinality 0,1,2,...,k. If one takes into account that higher order conditional independencies require larger amounts of data to remain stable, superior stability of this step in Fr(k)CI becomes obvious. Furthermore, this step was subdivided into two substeps, B') and B"). The first substep corresponds to technique used by FCI - restriting candidate sets of potential d-separators to the so far established neighbourhood. This substep is followed by the full search over all nodes of V - but only for edges left by B' - this is in contrast to FCI which omits step B) of the original CI, and thus runs into the troublers described in [10].

Step D2) has been modified in that the term "not d-connected" of CI was substituted by reference to local constraints. In this way results of step B) are exploited more thoroughly and in step D) no more reference is made to original body of data (which clearly accelerates the algorithm). This modification is legitimate since all the other cases covered by the concept of "not d-connected" of CI would have resulted in orientation of $D *-> B$ already in step C). Hence the generality of step D2) of the original CI algorithm is not needed here.

Steps D3) and D4) were interchanged as the step D3) of CI is quite time consuming and should be postponed until no alternative substep can do any-thing.

## 5 Properties of the Algorithm

Obviously, the algorithm Fr(k)CI will leave some edges actually not present in original data. As demonstrated in [11], superfluous edges may lead to incorrect belief network recovery. We shall show therefore that this is not the case with Fr(k)CI.

In [11] it has been proven that the original CI extended by abovementioned steps E) and F) will produce a dag compatible with the original data. Preliminaries for that result are that given the "real" dag G with visible variables $V_s$ and hidden ones $V_h$ one can define an "intrinsic" dag F in $V_s$ indistinguishable from G with respect to dependencies and independencies within set $V_s$ such that the modified CI algorithm produces a dag statistically indistinguishable from F. (This dag F is the IPG for G extended by removing every bidirectional edge $A <-> B$ and replacing it with a hidden node $H_{AB}$ adding edges $A < -H_{AB}- > B$). Below we show possibility of defining such an analogon of the dag F for the Fr(k)CI algorithm.

Let us define the r(k)-including path graph for G: G be a DAG with a set of hidden variables $V_h$ and of visible variables $V_s$. A graph $\pi$ be a r(k)-including path graph for G iff its set of nodes is $V_s$, and an edge between A and B from $V_s$ exists in $\pi$ iff no subset S of $V_s$ with cardinality not exceeding k does not d-separate nodes A and B in G. This edge is out of A iff there exists such a subset S' of $V_s$ with cardinality not exceeding k-1 that no trail in G from B to A into A is active with respect to S'. Otherwise this edge is ingoing into A.

**THEOREM 1** *Every edge in an r(k)-including path graph is either unidirectional or bidirectional (no edge is left unoriented)*

**PROOF:** Because there exists never a trail outgoing from A and outgoing from B which is active with respect to an empty set S ($card(S) = 0 \leq k$). Q.e.d.□

**THEOREM 2** *Let $\pi$ be r(k)-including path graph for G. If there is an edge $A->B$ in $\pi$, then there exists a directed path from A to B in G.*

**PROOF:** This is easily seen: Let S' be a subset of $V_s$ with cardinality not exceeding k-1 that no path in G from B to A into A is active with respect to S'. (1) Then clearly there must exist a trail in G outgoing out of A towards B which is active with respect to S' (otherwise edge AB would be absent from $\pi$ as S' would d-separate A and B). (2) Let us go along this trail in G as long as edges along it passing edges from tail to head. In this way we either reach B (which would complete the proof) or stop at a collider along this trail. This collider must either be in S' or have a successor in S' (as active trail definition requires). Let us continue the journey towards the blocking node in S'. The node is either not necessary for S' to block all ingoing trails from B to A (in this case we remove it from S' and start the procedure from the beginning - that is from point (1)) or it is necessary for that purpose. (3) In the latter case there is a trail between B and A ingoing into A this node is blocking. Let us continue our journey along this trail now in the direction where we pass edges from tail to head (at least one such direction exists). We continue at point (2). As the graph is a dag and the set S' is finite, the procedure is granted to terminate on reaching node B. This proves our claim. Q.e.d.□

**THEOREM 3** *Let $\pi$ be r(k)-including path graph for G. If we have two edges $A->B<-C$ with A and C not adjacent in $\pi$ then no subset S of $V_s$ with cardinality not greater than k containing B such that S d-separates A and C in G.*

**PROOF:** Because if such a set S existed then the set S-{B} with cardinality not greater than k-1 would have to block in G all trails from A to B into B or all trails from C to B into B (as this is required by definition of d-separation). But then the aforementioned definition of $\pi$ would require that either edge BA or BC resp. would be out of B. Q.e.d.□

**THEOREM 4** *Let $\pi$ be r(k)-including path graph for G. if A,B are adjacent in $\pi$, B,C are adjacent in $\pi$, but A,C are not adjacent in $\pi$ and on the trail A-B-C in $\pi$ node B is non-collider then there exists no such subset S of $V_s$* *with cardinality not greater than k not containing B that S d-separates A and C in G.*

**PROOF:** Otherwise if such a set S existed then (without restriction of generality let us assume $A < -B$) there exists a directed path from A to B in G. The set S would either block it or not. If not, then S would have to block all the trails from C to B which is a contradiction because then edge BC could not exist in $\pi$. Hence it must block it. But then S would have to block every trail ingoing into B either from direction of A or of C. Should it block those from direction of A (C) then there would exist an active trail outgoing from B towards A (C) and an active trail between B and C (A). But this is a contradiction as then there would exist an active trail connecting A and C (via B). This proves our claim. Q.e.d.□

**THEOREM 5** *Let $\pi$ be r(k)-including path graph for G. If there exists in $\pi$ a bidirectional edge between A and B, and if there exists an oriented path from A to B in G, and if there exists edge $C * - > A$ in $\pi$, then in $\pi$ there exists also the edge $C * - > B$.*

**PROOF:** As shown previously, $C * - > A < - > B$ means that there exists no set S containing A with cardinality k or lower such that C and B are d-separated in G. So let us consider sets S not containing A. Obviously, for every such set S there exists in G an active trail between C to A. If it is outgoing from A, then - as there must exist also an active trail between A and B, S will not d-separate A and B. So let us assume that S blocks all the trails outgoing from A to C. Then there must exist an active one from C to A ingoing into A. If the directed path from A to B in G is not blocked by S, then S does not d-separate C and B in G. Otherwise, if this path is blocked, there may exist an active trail from A to B outgoing from A, in which case S does not d-separate C and B in G either. If such a trail also does not exist, then there must exist an active trail from B to A ingoing into A. So let us combine these active trails from C into A and from B into A. This combined trail is also active as with respect to S because by assumption a successor of A belongs to S. Hence S does not d-separate C and B in G also. Hence, as in no case a set S of cardinality of k or below d-separates C and B in S, then there must exist an edge $C *-* B$ in $\pi$.

As we have orientation $C *-> A$ and $A <-> B$ then for every set S with cardinality k-1 or below there exist an active trail from C into A and from A into B in G. The latter is either out of A, in which can combination of both results in an active trail from C to B, or - if no active trail from A to B out of A exists - a successor of A is in S, and hence the combined trail is also active. So in any case, there exists an active trail from C to B into B in G, so the edge $C *-* B$ in $\pi$ must be oriented $C *-> B$.

Q.e.d.□

**Definition 1** *Let $\pi$ be r(k)-including path graph. Let FHG (full hiding*

*graph) of $\pi$ be a graph obtained from $\pi$ by preserving unidirectional edges and removing bidirectional edges replacing every bidirectional edge $A <-> B$ in $\pi$ with unidirectional edges $A <-H_{AB}-> B$, with H being a parentless hidden variable and A and B being not adjacent in the graph FHD.*

**THEOREM 6** *Let F be an FHG of a r(k)-including path graph $\pi$ of a DAG G. Then F is a DAG.*

**PROOF:** It is immediately visible that every directed path in $\pi$ preserves node ordering from G. Since F keeps these directed paths and adds only parentless nodes, the ordering of nodes imposed by G is preserved in F. Q.e.d.□

**THEOREM 7** *Let F be an FHG of a r(k)-including path graph $\pi$ of a DAG G. Let S be a subset of the set of all visible nodes of G. Let $card(S) \leq k$. Then S d-separates visible nodes X and Y in G (both not in S) if and only if S d-separates them in F.*

**PROOF:** To show this, one needs only to demonstrate that an active trail in G is also active in F and an active trail in F is also active in G.

Part I: First let us consider an active trail in F. Let it be minimal that is for every three successive nodes A,B,C ($A *-* B *-* C$) on this trail A and B are not adjacent in F (Otherwise, as demonstrated in [9], a minimal active trail can always be derived from it, and we can therefore always consider a minimal active trail, if an active trail exists). Let B be a non-collider on this trail that is $A<-B->C$ or $A<-B<-*C$ or $A*->B->C$ on that trail. As B does not belong to S, there must always exist an active trail in G with respect to S connecting A and C (as otherwise B would have to be a collider). Let us consider now a collider B. B must then be visible in F, and in $\pi$ we have $A*->B<-*C$. If S in G either does not block an active trail out of B to A or out of B to C, then as previously there exists an active trail in G connecting A and C via B. Otherwise there is an active trail in G from A into B and from C into B. We recall the fact that B has in F a successor or is itself in S, hence it has in G the same successor in S or is in S. Hence the combined trail from A into B and from C into B is an active trail from A via B to C in G. By induction we come to the conclusion that a (minimal hence every) active trail in F has an active counterpart in G. Hence non-d-separation in F implies non-d-separation in G.

Part II: Let us consider an active trail in G. First, let us consider two close visible nodes on this trail. If there are no nodes between them or no hidden node between them is a collider, then in $\pi$ they are neighbours. Otherwise, there is a trail between them in $\pi$ such that every node on this trail is a collider and has a successor or is itself in S. Hence this (sub)trail is active in F. Let us consider three close visible nodes in G on the active trail A,B,C. If B is a non-collider in G and non-collider in F, then active subtrails AB and BC extend to an active subtrail AC in F. If B is non-collider in G, but it is a collider in F, then there exists an edge between A and C in $\pi$, hence an active trail between them in F. If B is a collider in G, then it is a collider in $\pi$. Now if it has a successor D in G in S (or is itself in S) and also in F in S, then there exists an active trail between A and C in F. It may, however, happen that an edge on the path from B to D in G turns to a bidirectional one in $\pi$, hence D may not be a successor of B in F along the original path. But as visible from previous theorems, a shortened path will substitute it so that D still remains a successor of B or in the worst case we have $A * -> D < -* C$ in $\pi$, hence we have still an active trail between A and C in F. By induction we can extend the active trail in F to both ends X,Y of the active trail from G. Hence non-d-separation in G implies non-d-separation in F.

Q.e.d.□

**THEOREM 8** *Let F be an FHG of a r(k)-including path graph $\pi$ of a DAG G. Let S be a subset of the set of all visible nodes of G. (Let card(S) be unrestricted). If S d-separates visible nodes X and Y in F (both not in S) then S d-separates them in G.*

**PROOF:** To prove this, notice that in the second part of the proof of the preceding theorem no restriction on cardinality of S was required.

Q.e.d.□

This means that a bidirectional edge $A <-> B$ in $\pi$ can be treated as a unidirectional edges $A<-H->B$, with H being a parentless hidden variable and A and B being not adjacent in the graph.

The aforementioned statements indicate that for a faithful graph G for edge pair A-B and B-C with A and C not adjacent a statistical test of independence of A and C relatively to sets S containing B with cardinality not greater than k will correctly decide about orientation of edges with respect to the r(k)-including path graph $\pi$. Clearly, as in case of including path graphs, r(k)-including path graphs cannot be fully recovered from data. What we produce in steps A)-D) of the above algorithm, is a partial r(k)-including path graph. A partial r(k)-including path graph differs from the r(k)-including path graph in that it contains non-oriented edge ends $o$, e.g. $Ao->B$, and some local constraint information e.g. $A*-\underline{*B*}-*C$ meaning that edges AB, BC cannot meet head to head at B.

With these prerequisites let us present a correctness proof of Fr(k)CI algorithm.

**THEOREM 9** *Step B'-B" produces a partial r(k)-including path graph having only edges between those nodes where are edges in the intrinsic r(k)-including path graph*

To prove this, one needs only to compare these steps to the definition of r(k)-including path graph.

**THEOREM 10** *Steps C and D orient edges identically with and produce local constraints consistent with the intrinsic r(k)-including path graph*

This is the consequence of earlier theorems on relationship between G and r(k)-including path graph.

**THEOREM 11** *(i) Steps E and F produce a belief network (ii) keeping all dependencies and independencies of the intrinsic FHG.*

See Appendix A.

**THEOREM 12** *The algorithm produce a belief network keeping all dependencies and independencies of the intrinsic underlying DAG G for conditioning sets of cardinalities up to k and all independencies indicated by this belief network are also present in the intrinsic underlying DAG G.*

This is the direct consequence of the previous theorem and Theorem 7 and Theorem 8.

# 6  Experiments

Two types of experiments have been carried out: first type assuming "perfect statistical tests" that is with conditional independence tests being answered based on d-separation within the intrinsic underlying DAG, and the second one with simulated random sample from the intrinsic underlying DAG.

The first type of tests served as a kind of feasibility study. The largest network tested was the so-called ALARM-network ( 37 nodes, 46 edges, see [3] page 330 for its scheme). The original CI algorithm [20] had no chance to terminate on my PC-AT. The original FCI algorithm [20] required less than 9,000 "conditional independence tests" to complete (producing correct partial including path graph structure). The Fr(1)CI required on the other hand less than 10,000 "tests", Fr(2)CI algorithm - about 40,000 "tests". Fr(1)CI produced a network with 60 superfluous edges, and Fr(2)CI with 1 superfluous edge. The original "underlying distribution" was reconstructed correctly. Other experiments with ALARM network consisted in making some randomly selected nodes "hidden ". In these cases also the number of statistical tests required had approximately same proportions. It seems that ensuring that the recovered belief network structure is really correct is quite expensive.

For smaller networks, 10-20 nodes, up to 30 edges, both types of experiments have been carried out. It seems that application of Fr(k)CI with k up to 3, satisfactorily recovers the underlying belief network structure and distribution

# 7  Discussion and Concluding Remarks

Within this paper a new algorithm of recovery of belief network structure from data has been presented and its correctness demonstrated. It relies essentially on "acceleration" of the known CI algorithm of Spirtes, Glymour and Scheines [20] by restricting the number of conditional dependencies checked up to k variables and it extends CI by additional steps transforming so called partial including path graph into a belief network. Sample outputs of CI, Fr(1)CI and Fr(2)CI are shown in Fig.1. Though Fr(k)CI introduces redundant edges (e.g. AC and FC in Fig.1b), indicating dependencies not present in the original data, it actually avoids pitfalls of the FCI algorithm, another CI "accelerator" proposed by Spirtes, Glymour and Scheines [20], as visible from section 5 and [11].

Undoubtedly, relationship between CI, FCI, Fr(k)CI (not discussed in detail due to space limitations) raises the question of shades of correctness. Given a fully reliable statistical test, CI provides with absolutely correct structure of distribution. FCI delivers exactly the same structure as CI but in FCI-unfriendly cases (special structures in about 30 variables and more) when erroneously non-existent independencies are indicated and statistically inconsistent causal relationship is indicated. Fr(k)CI on the other hand approaches CI output in a consistent manner in that all independencies indicated by Fr(k)CI are correct (though some existent independencies may not be discovered) and some indicated direct causalities may be in fact indirect ones. The only factor balancing merits seems then to be the speed, with FCI and Fr(k)CI usually outperforming CI. (Speed comparison between FCI and Fr(k)CI is a function of intrinsic problem structure). However, as CI requires usually much more statistical testing than the other two, it runs a greater risk of statistical error for a given sample size. FCI requires usually higher order conditional independence tests than Fr(k)CI and is at this end more vulnerable, but it may (for a given intrinsic problem structure) require fewer statistical test than Fr(k)CI.

[21] T.S. Verma, J. Pearl: *Equivalence and synthesis of causal models*, Proc. of the Conference on Uncertainty in Artificial Intelligence, Cambridge MA (1990), 220-227.

# Appendix A

**Proof of Theorem 11**

Let us consider a mixed graph (MG) having all the unoriented edges of the original r(k)-including path graph (here called FHD - Full Hiding Dag) and the Fr(k)CI output after stage D (here called Fr(k)CI-AD graph). Let all the unidirected edges of FHD be unidirected the same way in MG, let all the edges bidirected by Fr(k)CI-AD he bidirected in MG. Let all partially directed edges of Fr(k)CI-AD be unidirected in MG. Last not least, let all bidirectional FHD edges not oriented at all by Fr(k)CI-AD be left unoriented in MG. Now if there were no cycles in MG, then also the claim (i) of the above theorem would be valid. The proof of acyclicity of MG is not difficult, but laborious. An overview can be made in terms of Figures, from Fig.3 to Fig.12. In this series of Figures it is demonstrated that no three edges of MG can form a cycle.

Figures summarize the proof as follows: In Fig.3 three possibilities of triangles $\Delta 1$ (with one bidirectional edge), $\Delta 2$ (with 2 bidirectional edges), $\Delta 3$ (with 3 bidirectional edges) in the intrinsic r(k)-including path graph,

which give risk of cycles in MG, are shown. Figures 4-5 review possible MG situations for $\Delta 1$, Figures 6-10 - for $\Delta 2$, and Figures 11-12 - for $\Delta 3$.

Within the Fr(k)CI-AD algorithm, only orientation steps denoted as $D_p$, $D_s$ and $C$ can give rise to a partial orientation of a FHD-bidirectional edge within Fr(k)CI-AD. (Step $D_d$ immediately creates a bidirectional edge in Fr(k)CI-AD out of it). Therefore each type of triangle is considered for potential causes of cycles due to these steps of the algorithm. Within case $\Delta 1$ each time influence of one of these steps is considered, within case $\Delta 2$ - combination of two such steps, and in case $\Delta 3$ - combination of three steps is taken into account.

E.g. Fig.5 (a) overviews the general situation in the FHD when one edge (BA) is bidirectional and AC and CB are unidirectional and the edge DA causes in step ($C$) of Fr(k)CI-AD orientation of BA and DA towards A. (b) and (c) follow the case when the edges DA and AC are not bridged (that is their non-common ends do not share any other edge): as a result the edge BA is made bidirectional in Fr(k)CI-AD. On the other hand, (d) and (f) deal with the case when DA and AC are bridged: then also BA grows bidirectional in Fr(k)CI-AD. So. Fig.4-5 show that in case of one FHD-bidirectional and two FHD-unidirectional edges no cycle in MG is possible. Fig.6-10 demonstrate the same for two FHD-bi- and one FHD unidirectional edges, demonstrating, that both bidirectional edges will be made bidirectional in MG if there were any risk of cyclicity during the Fr(k)CI-AD-algorithm. Fig.11-12 are concerned with potential cycles consisting of three FHD-bidirectional edges

FAST RESTRICTED CAUSAL INFERENCE                    29FAST RESTRICTED CAUSAL INFERENCE 29FAST RESTRICTED CAUSAL INFERENCE    29

Within a longer trail of edges it is immediately visible, that there must exist at least one pair of neighboring edges (one of them bidirectional in FHD) which are "bridged" that is their ends not neighboring on the path are neighbors in the graph. Hence we have here a triangle which - due to facts proven earlier - cannot form a cycle in MG and at least two edges and at least one FHD-bidirectional are oriented correctly (that is as in FHD), hence cannot participate also in a larger cycle. (This is clear if the "third" edge has also been oriented by $C$, $D_s$ or $D_p$ step. But we can easily check, that if it has been oriented by the $D_d$ step, then the other edges will be oriented prohibiting a long-run cycle.) This completes the proof of claim (i).

As claim (ii) is concerned, we shall first notice that a situation like that of Fig.13 cannot happen in an including path graph, that is it is never possible, that along a path $AB_1...B_nC$ with head to head meetings at $B_i$ one edge outgoing from each $B_i$ points at A and there is some j such that an edge $B_jC$ is outgoing from $B_j$. Now, when orienting edges according to Fr(k)CI-EF algorithm, we can make two types of errors:(a) introduce a path which is not active (in terminology of [4]) in BN, but is actually active in FHD, and (b) introduce a path which is active (in terminology of [4]) in BN, but is actually not active (blocked) in FHD. In case (a), we may have the structure of such a path as $..., D, B_n, ..., B_1, A, C_1, ..., C_m, E, ...$ in Fig. 14, with node A set in BN erroneously active (to the left) or passive (to the right). Let us assume that this is the shortest active path between the nodes of interest that is no subset of nodes on the erroneously active path can form also an active path. Then in Fig.14.a) and .b) there exists no unioriented edge $D->B_i$ nor $E->C_j$, nor bidirectional edge $B_i <->C_j$ nor $D<->B_j$ nor $E<->C_j$ nor $D<->A$ nor $E<->A$, for any i,j. And additionally in Fig.14.a) there exists no edge $D->A$ nor $E->A$. In Fig.14.b) there exists no edge $D<-A$ nor $E<-A$. But it can then be demonstrated, that the Fr(k)CI-AD orients correctly nodes from D to $B_1$ and from E to $C_1$, and then a definite discriminating path for A emerges, and the edges at A are oriented correctly, hence it is denied that an error may occur at A.

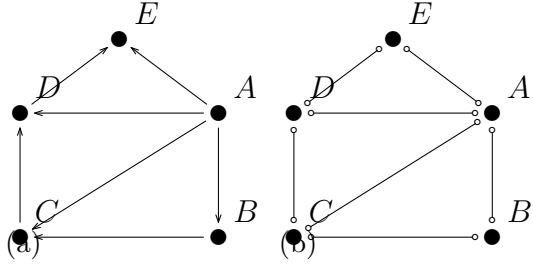

Figure 1: A result of running CI algorithm: (a) the original dag (b) the dag recovered

As error (b) is concerned, we can proceed in an analogous way, also assuming that we have to do with the shortest erroneously passive path.

This would then complete the proof of the Theorem.

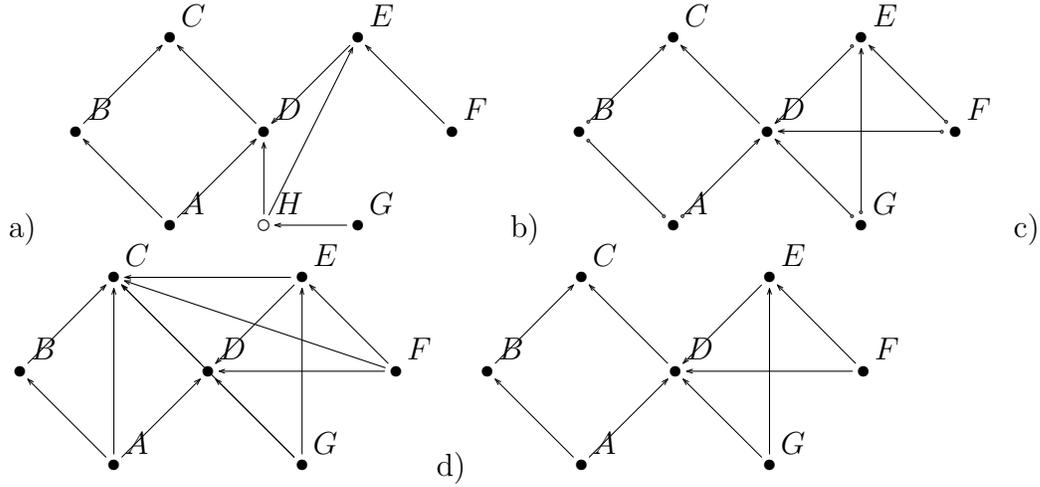

Figure 2: a) The underlying DAG G, b) as recovered by original CI algorithm, c) as recovered by Fr(1)CI, d) as recovered by Fr(2)CI.

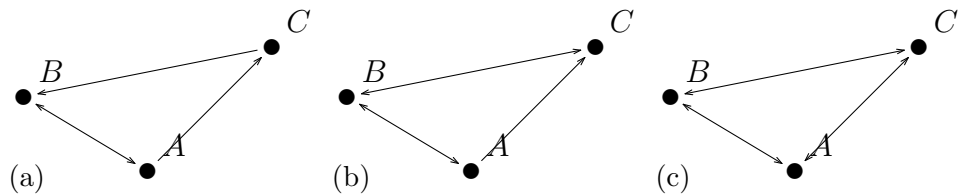

Figure 3: $\Delta$ - a triangle in the original r(k)-including path graph.(a)- $\Delta 1$ - only one intrinsic bidirectional edge(b)- $\Delta 2$ - only two intrinsic bidirectional edges(c)- $\Delta 3$ - three bidirected edges

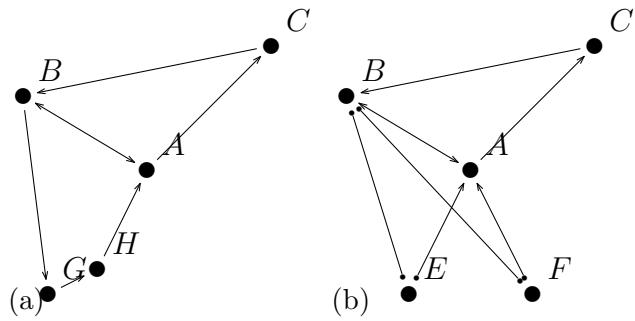

Figure 4: (a)- $\Delta 1 D_p$ - orientation due to a directed path - impossible as introducing a cycle(b)- $\Delta 1 D_s$ - orientation due to a separation - impossible as oriented path $A-> C-> B$ renders E,F dependent given B (denying defining condition of $(D_s)$ for arrow $B*->A$)

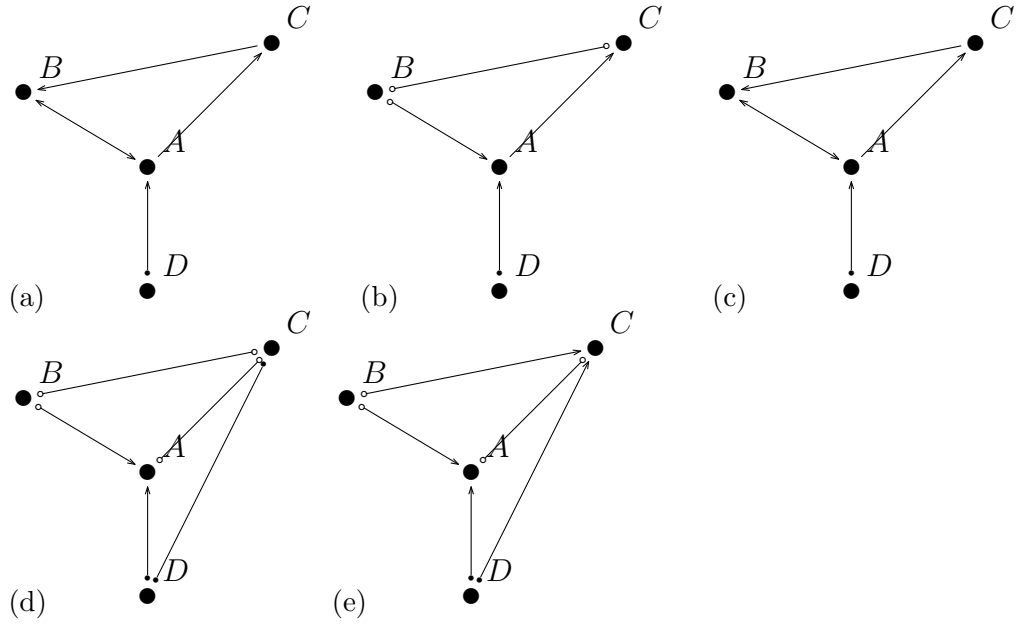

Figure 5: (a)- $\Delta 1C$ - orientation due to unbridged head to head(b)- $\Delta 1C(1)$ - case 1 (Fr(k)CI-AD) - due to $(C)$ on path DAC(c)- $\Delta 1C(1)b$ - hence due to discriminating path DACB for C and DABC for B - O.K.(d)- $\Delta 1C(2)$ - case 2 (Fr(k)CI-AD) - when DAC is bridged(e)- $\Delta 1C(2)b$ - hence due to $(C)$ on DCBand due to A being a collider along DAB and the edge $A->C$- but this is impossible as it denies the intrinsic structure for edge $C->B$

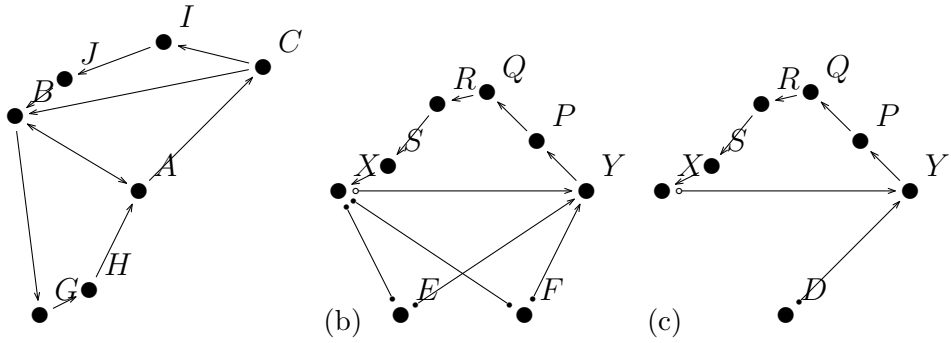

Figure 6: (a)- $\Delta 2 D_p D_p$ - impossible -due to a cycle(b)- $\Delta 2 D_p D_s$ - impossible -due denying orientation capability of $X *-> Y$(c)- $\Delta 2 D_p C$ - impossible -due to denying non-connection of X and D

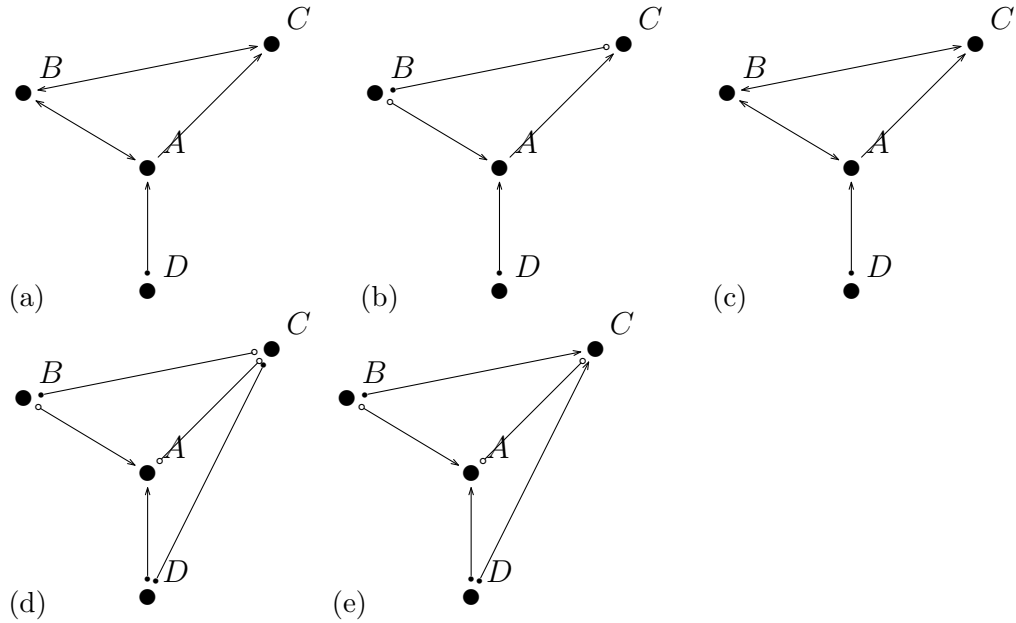

Figure 7: (a)- $\Delta 2C@A$ - case $(C)$-orientation applied at A(b)- $\Delta 2C@A(1)$ - case DAC not bridged(c)- $\Delta 2C@A(1)b$ - hence, due to discriminating path DACB for C and DABC for B - O.K.(d)- $\Delta 1C@A(2)$ - case 2 (Fr(k)CI-AD) - when DAC is bridged(e)- $\Delta 1C@A(2)b$ - hence due to $(C)$ on DCBand due to A being a collider along DAB and the edge $A->C$- O.K.

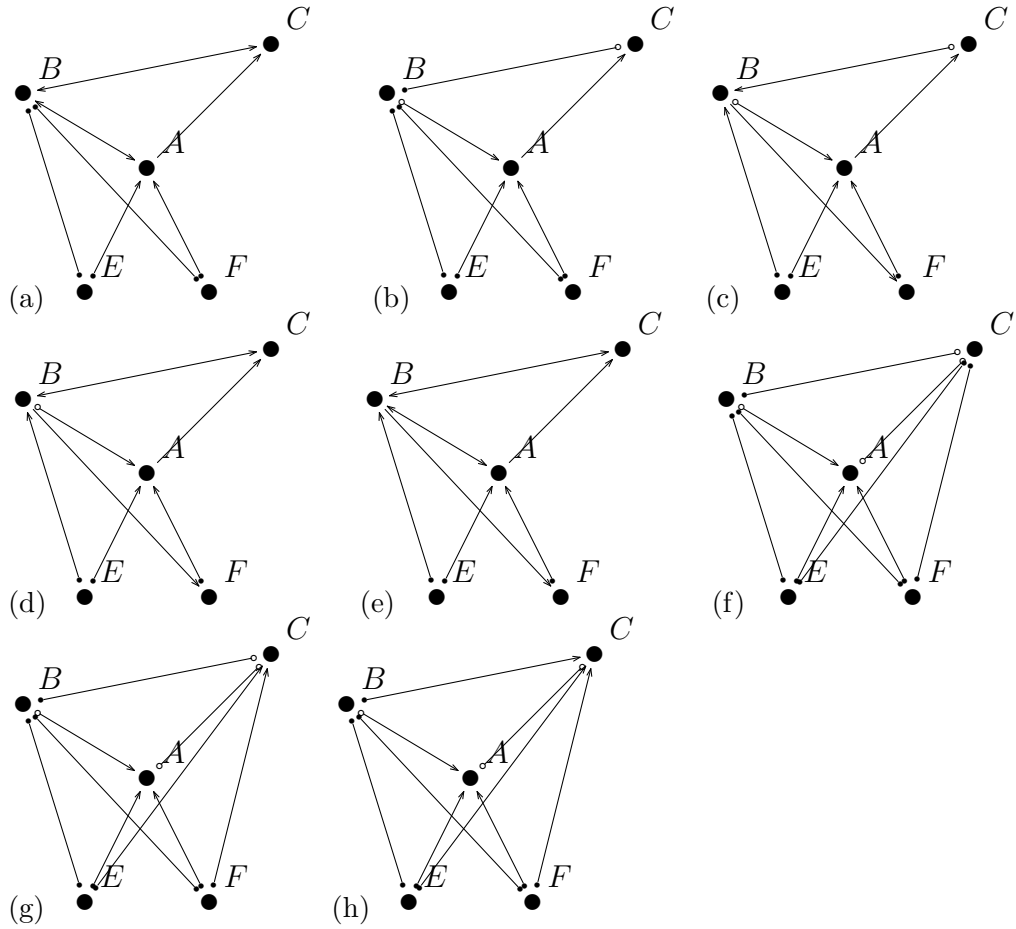

Figure 8: (a)- $\Delta 2D_s@A$ - case $(D_s)$-orientation applied at A(b)- $\Delta 2D_s@A(1)$ - case EAC (or by analogy: FAC) not bridged(c)- $\Delta 2D_s@A(1)b$ - hence due to non-bridged path EBC and EBF (d)- $\Delta 2D_s@A(1)c$ - hence due to definite discriminating path EACB for C(e)- $\Delta 2D_s@A(1)d$ - hence due to definite discriminating path EABC for B - O.K.(f)- $\Delta 2D_s@A(2)$ - case both EAC and FAC bridged(g)- $\Delta 2D_s@A(2)b$ - hence due to unbridged ECF(h)- $\Delta 2D_s@A(2)c$ - hence due to d-separability of E,F given B - O.K.

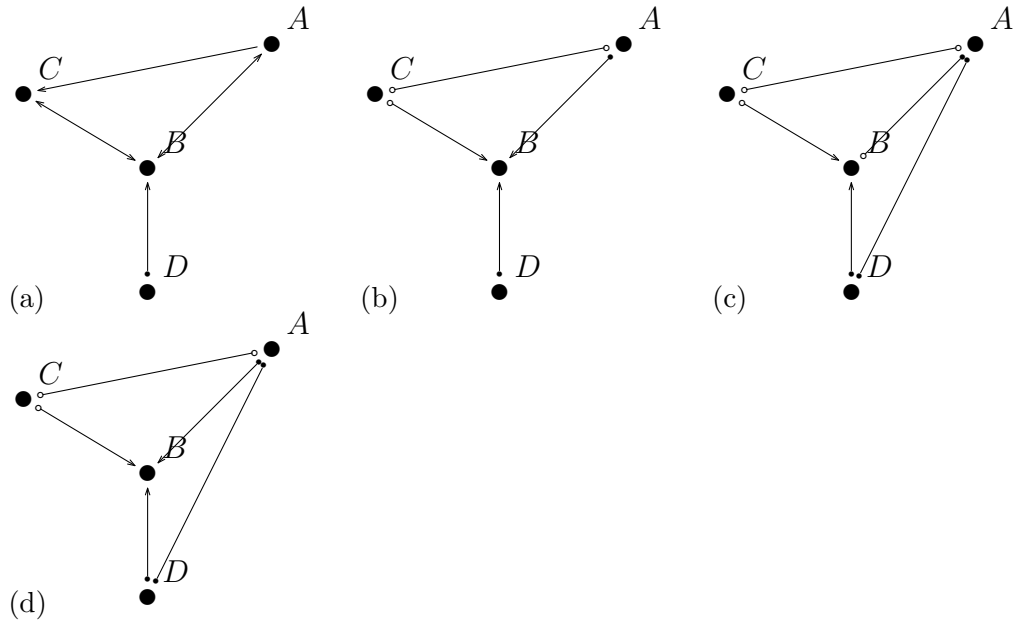

Figure 9: (a)- $\Delta 2C@B$ - case ($C$)-orientation applied at B(b)- $\Delta 2C@B(1)$ - case DAC not bridged - O.K.(c)- $\Delta 2C@B(2)$ - case DAC bridged(d)- $\Delta 2C@B(2)b$ - due to ($D_s$ at A for collider $C*->B<-*A$)- O.K.

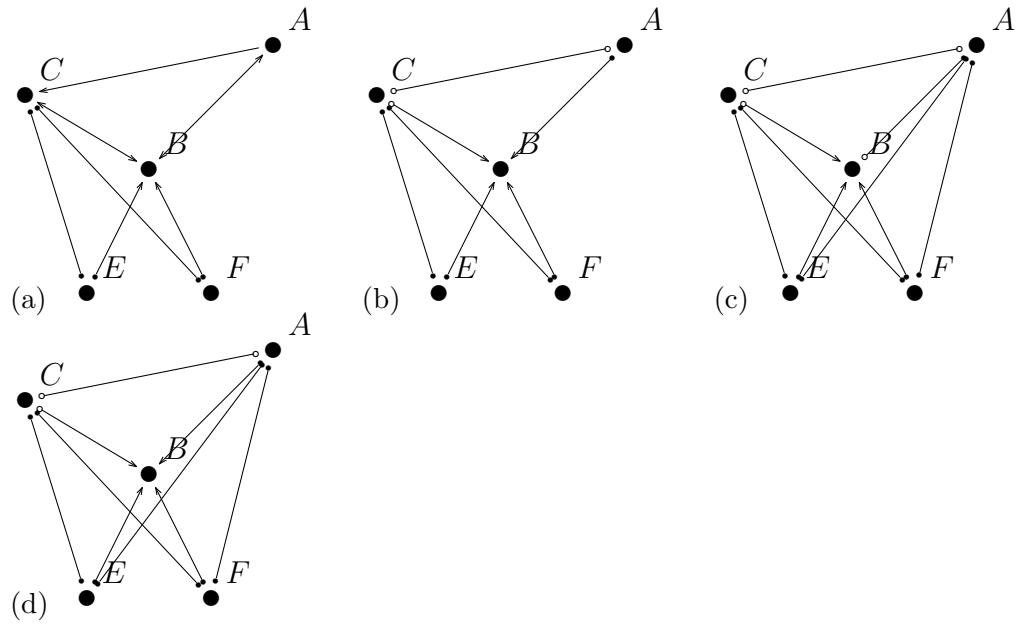

Figure 10: (a)- $\Delta 2D_s@B$ - case $(D_s)$-orientation applied at B(b)- $\Delta 2D_s@B(1)$ - EBA (or by analogy: FBA) not bridged- O.K.(c)- $\Delta 2D_s@B(2)$ - EBA and FBA bridged(d)- $\Delta 2D_s@B(2)b$ - due to $(D_s$ at A for collider $E*->B<-*F$O.K.

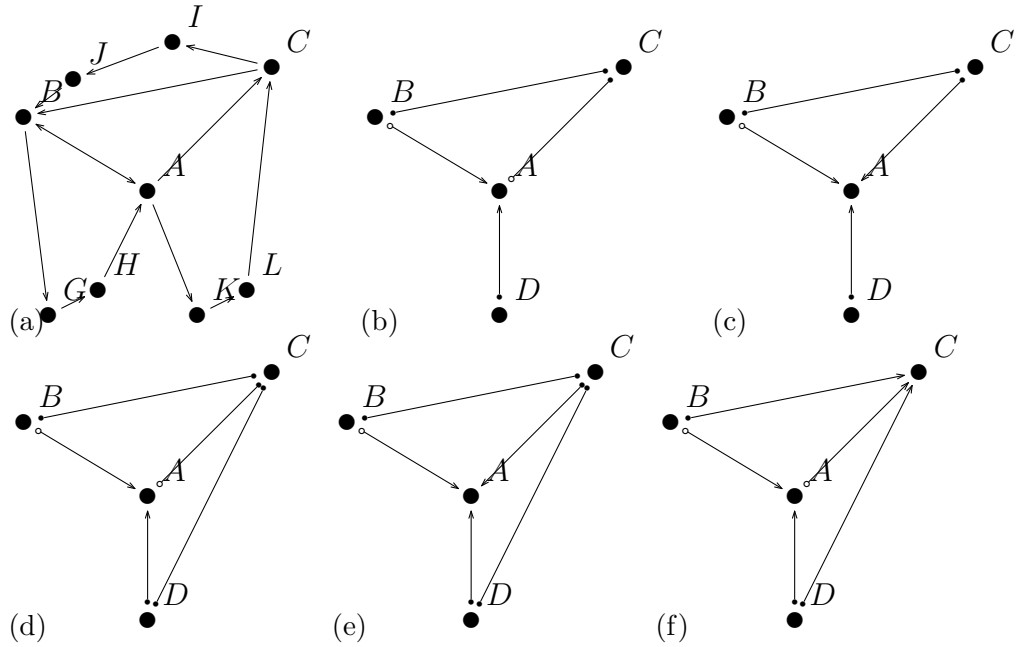

Figure 11: (a)- $\Delta 3 D_p D_p D_p$ - impossible -due to a cycle(b)- $\Delta 3C$ - full symmetry in all nodes A,B,C(c)- $\Delta 3C(1)$ - DAC unbridged - O.K.(d)- $\Delta 3C(2)$ - DAC bridged(e)- $\Delta 3C(2)b(1)$ - if $(D_s)$ at node C for DAB applicable- O.K.(f)- $\Delta 3C(2)b(2)$ - if $(D_s)$ at node C for DAB not applicable - O.K. ($D_p$ for $C *-> B$ impossible as non-connection of D and B would be denied.

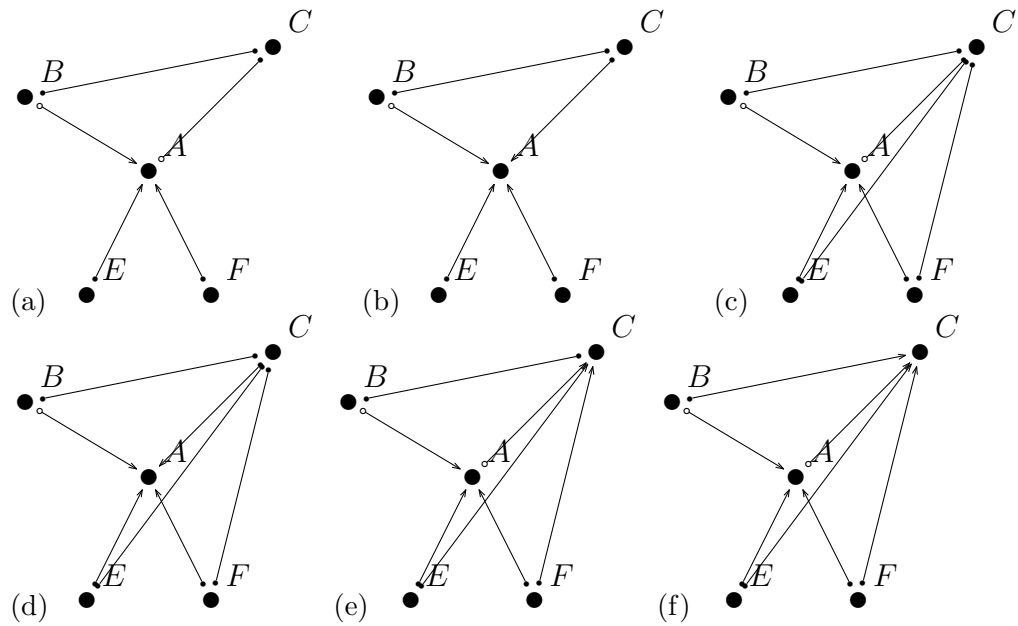

Figure 12: (a)- $\Delta 3D_s$ - full symmetry in all nodes A,B,C(b)- $\Delta 3D_s(1)$ - EAC unbridged - O.K.(c)- $\Delta 3D_s(2)$ - FAC and EAC bridged(d)- $\Delta 3D_s(2)b(1)$ - if $(D_s)$ at node C for EAF applicable- O.K.(e)- $\Delta 3D_s(2)b(2)$ - if $(D_s)$ at node C for EAF not applicable,(f)- $\Delta 3D_s(2)b(2)b$ - then $(D_s$ at node B for ECF applicable, $(D_p$ for $C*->B$ would then not be appropriate)

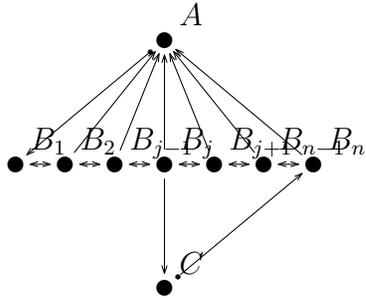

Figure 13: Impossible because $B_j$ cannot participate in d-separation of A and C

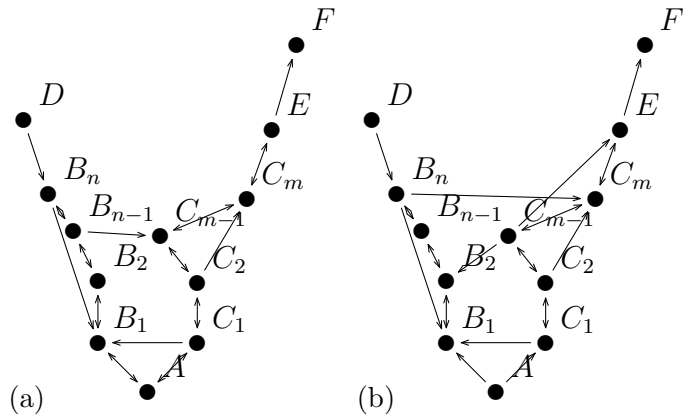

Figure 14: hdags possibly causing node A to be erroneously considered as (a) non-collider (b) a collider.